\documentclass[sigconf,]{acmart}
\setcopyright{none}
\renewcommand\footnotetextcopyrightpermission[1]{}

\usepackage{enumitem} 
\usepackage{xspace}
\usepackage{graphicx}
\usepackage{tabularx}
\usepackage{multirow}
\usepackage{soul}
\usepackage{hyperref}
\usepackage{subcaption}
\usepackage{booktabs}
\usepackage{listings}
\usepackage{xcolor} 
\definecolor{darkgreen}{RGB}{0,128,0}
\newcommand{\ours}{{PyOD 2}\xspace}

\AtBeginDocument{%
  }

\setcopyright{acmlicensed}
\copyrightyear{2018}
\acmYear{2018}
\acmDOI{XXXXXXX.XXXXXXX}

\acmConference[Conference acronym 'XX]{Make sure to enter the correct
  conference title from your rights confirmation emai}{June 03--05,
  2018}{Woodstock, NY}
\acmISBN{978-1-4503-XXXX-X/18/06}

\begin{document}


\title{\ours: A Python Library for Outlier Detection with LLM-powered Model Selection}



\author{Sihan Chen\textsuperscript{1,*}, Zhuangzhuang Qian\textsuperscript{1,*}, Wingchun Siu\textsuperscript{1,*}, Xingcan Hu\textsuperscript{1}, Jiaqi Li\textsuperscript{1}, Shawn Li\textsuperscript{1}, \\ Yuehan Qin\textsuperscript{1}, Tiankai Yang\textsuperscript{1}, Zhuo Xiao\textsuperscript{1}, Wanghao Ye\textsuperscript{2},  Yichi Zhang\textsuperscript{1}, Yushun Dong\textsuperscript{3}, Yue Zhao\textsuperscript{1}}
\affiliation{%
  \institution{\textsuperscript{1}University of Southern California, \textsuperscript{2}University of Maryland, \textsuperscript{3}Florida State University}
  \city{}
  \country{}
}
\email{
  {schen976, alexqian, siuw, xh_186, jli77629, li.li02, yuehanqi, tiankaiy, zhuoxiao, yzhang42, yzhao010}@usc.edu,
}
\email{wy891@umd.edu, ydong@fsu.edu}
\thanks{*Equal contribution. The remaining authors are listed alphabetically by last names.}

\renewcommand{\shortauthors}{Trovato et al.}

\begin{abstract}


Outlier detection (OD), also known as anomaly detection, is a critical machine learning (ML) task with applications in fraud detection, network intrusion detection, clickstream analysis, recommendation systems, and social network moderation.
Among open-source libraries for outlier detection, the Python Outlier Detection (\textbf{PyOD}) library is the most widely adopted, with over 8,500 GitHub stars, 25 million downloads, and diverse industry usage. 
However, PyOD currently faces three limitations: (1) insufficient coverage of modern deep learning algorithms, (2) fragmented implementations across PyTorch and TensorFlow, and (3) no automated model selection, making it hard for non-experts.

To address these issues, we present \textbf{PyOD Version 2 (\ours)}, which integrates 12 state-of-the-art deep learning models into a unified PyTorch framework and introduces a large language model (LLM)-based pipeline for automated OD model selection. 
These improvements simplify OD workflows, provide access to 45 algorithms, and deliver robust performance on various datasets. 
In this paper, we demonstrate how \ours streamlines the deployment and automation of OD models and sets a new standard in both research and industry.
\ours is accessible at \url{https://github.com/yzhao062/pyod}.

This study aligns with the \textbf{Web Mining and Content Analysis} track, addressing topics such as the \textit{robustness of Web mining methods} and the \textit{quality of algorithmically-generated Web data}.

\end{abstract}




\maketitle

\section{Introduction}


Outlier detection (OD), also known as anomaly detection (AD),\footnote{We use both terms interchangeably in this paper.} is a central task in data analysis and machine learning \cite{han2022adbench,liu2022bond}. It has critical applications in web-related domains, including fraud detection in e-commerce \cite{lee2020autoaudit}, clickstream analysis for user behavior modeling \cite{beutel2015graph}, software engineering \cite{Sunicse}, and social network moderation \cite{Sunicse}.
While traditional methods have been extensively studied and applied \cite{aggarwal2015outlier}, deep learning methods have recently gained attention. Models based on autoencoders \cite{aggarwal2015outlier}, generative adversarial networks (GANs) \cite{goodfellow2020generative}, mixture-of-experts \cite{zhao2023admoe}, and transformers \cite{vaswani2017attention} have shown strong performance in detecting anomalies within complex, high-dimensional data \cite{pang2021deep,han2022adbench,li2024dpu, Li_Ji_Wu_Li_Qin_Wei_Zimmermann_2024,Limm}.

\noindent
\textbf{Current Landscape of Open-source OD Systems}.
Among the open-source libraries available for outlier and anomaly detection, PyOD \cite{zhao2019pyod} is not only the most widely used one, with more than 8,500 GitHub stars, 25 million downloads, and more than 1,000 citations, but it has also become a trusted resource in both academic and industrial communities. 
Its broad coverage of algorithms, ranging from classical statistical techniques to more advanced machine learning models, has earned it a strong following among developers, analysts, and researchers who rely on OD capabilities as a core component of their workflows.
Despite its established position and extensive user base, PyOD still faces three key challenges: (\textit{i}) a lack of modern deep learning methods, (\textit{ii}) the coexistence of two different frameworks (PyTorch and TensorFlow) for neural models, and (\textit{iii}) the absence of automated model selection. 
These limitations create integration issues, reduce consistency, and place a heavy burden on the domain expertise of outlier detection, thereby limiting its accessibility for less experienced users.


We address these challenges by introducing \textbf{PyOD Version 2 (\ours)}.
As described in \S \ref{sec:design}, \ours provides:
\begin{enumerate}[leftmargin=*]
    \item \textbf{Expanded Deep Learning Support}: \ours integrates 12 state-of-the-art neural models, refactored into a single PyTorch framework, bringing the total number of OD models to 45.
    \item \textbf{Enhanced Performance and Ease of Use}: All models are optimized for efficiency and consistency, ensuring reliable performance across various datasets.
    \item \textbf{First LLM-driven OD Model Selection}: \ours introduces an automated model selection mechanism, using an LLM-based reasoning process. This reduces the need for manual tuning and supports users who may lack deep domain knowledge.
\end{enumerate}


\noindent
\textbf{Demonstration}.
In \S \ref{sec:demo}, we demonstrate (1) how to use \ours to integrate the latest OD models such as LUNAR \cite{goodge2022lunar} into workflows in just five lines of code and (2) show how its LLM-based model selection provides effective guidance for choosing OD models from a pool of 10 models on 17 OD datasets.

\begin{figure*}[h]
    \centering
    \includegraphics[width=1\linewidth]{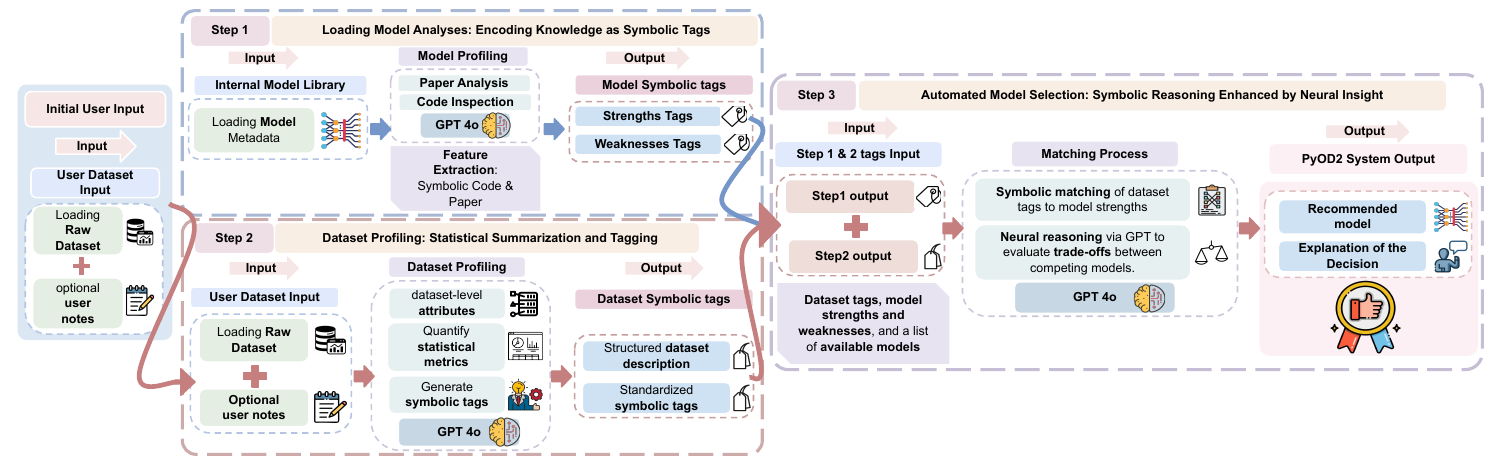}
    \caption{An overview of the automated three-step pipeline for model selection in \ours (see \S \ref{subsec:llm}). 
Step 1 analyzes each model's paper and code to produce symbolic tags describing its strengths and weaknesses. 
Step 2 profiles the dataset, generates statistical summaries, and produces symbolic tags characterizing it. 
Step 3 uses these model and dataset tags, combined with LLM-based reasoning, to identify the most suitable model and provide an explanation of the decision.}
\vspace{-0.2in}
    \label{fig:overview}
\end{figure*}




\section{Overview and Design of \ours}
\label{sec:design}

The original PyOD \cite{zhao2019pyod} offers a unified API that covers a wide range of algorithms, including proximity-based methods, linear models, neural methods, and ensemble techniques. It provides JIT optimizations, parallelization, cross-platform compatibility, and extensive testing, making it robust and easy to integrate.

Building on these foundations, \ours introduces major improvements in deep learning algorithm support and usability (\S\ref{subsec:expansion}) while adding a novel model selection framework powered by LLMs (\S\ref{subsec:llm}). These updates modernize the framework and ensure \ours remains flexible, reliable, and adaptable.

\subsection{Enhanced Coverage and Usability}
\label{subsec:expansion}

\ours significantly strengthens its deep learning capabilities by integrating advanced algorithms and improving ease of use through a unified PyTorch-based design.

\noindent \textbf{Integration and Optimization of Deep Learning Models.}
\ours adds over ten state-of-the-art deep learning OD models (see Table \ref{table:algorithms}), including MO-GAAL \cite{liu2019generative}, SO-GAAL \cite{liu2019generative}, AE \cite{aggarwal2015outlier}, VAE \cite{kingma2013auto}, AnoGAN \cite{schlegl2017unsupervised}, DeepSVDD \cite{ruff2018deep}, ALAD \cite{zenati2018adversarially}, AE1SVM \cite{nguyen2019scalable}, DevNet \cite{pang2019deep}, and LUNAR \cite{goodge2022lunar}. 
Each model is optimized for unified usability, training stability, memory management, and performance.

\noindent \textbf{Unified PyTorch Framework.}
All neural-based models in \ours now run on a unified PyTorch framework \cite{paszke2019pytorch}, enabling consistent execution, simplified debugging, and efficient integration of future methods. The PyTorch environment also supports GPU acceleration, improving scalability and performance.

\noindent \textbf{Standardized Deep Learning Architecture.}
\ours introduces a base class, \texttt{base\_dl}, providing a standard structure for all deep learning models. Common operations, including model initialization, training, and evaluation, are standardized, reducing code redundancy and simplifying the addition of new models.

\begin{table*}[ht]
\centering
\caption{List of integrated deep learning-based outlier detection models in \ours.}

\label{tab:dl_models}
\scalebox{0.95}{
\begin{tabular}{@{}l|p{9cm}lcl@{}}
\toprule
\textbf{Model} & \textbf{Description} & \textbf{Backbone} & \textbf{Year} & \textbf{Reference} \\ 
\midrule
AutoEncoder (AE) & Encodes data into a compressed representation and detects outliers by measuring reconstruction errors in the output compared to the input. & AutoEncoder & 2015 & \cite{aggarwal2015outlier} \\ 
Variational AE (VAE) & Extends AE by learning probabilistic latent representations of data, detecting anomalies via reconstruction probabilities. & AutoEncoder & 2015 & \cite{kingma2013auto} \\ 
AnoGAN & Leverages GANs to model normal data distributions and identifies anomalies as deviations from the learned normal manifold. & GAN & 2017 & \cite{schlegl2017unsupervised} \\ 
DeepSVDD & Optimizes a hypersphere in the feature space to enclose normal data points, minimizing the volume and isolating outliers. & Neural Network & 2018 & \cite{ruff2018deep} \\ 
ALAD & Adopts bidirectional GANs to map data to latent space and back, detecting anomalies by comparing the alignment of data and latent mappings. & GAN & 2018 & \cite{zenati2018adversarially} \\ 
MO-GAAL & Uses multiple GAN generators to generate diverse synthetic outliers, enhancing the ability to identify challenging anomalies. & GAN & 2019 & \cite{liu2019generative} \\ 
SO-GAAL & Employs a single GAN generator to directly produce potential outliers, refining the boundary between normal and abnormal data. & GAN & 2019 & \cite{liu2019generative} \\ 
AE1SVM & Combines the reconstruction power of AE with the boundary-based classification of One-Class SVM for improved anomaly detection. & AutoEncoder + SVM & 2019 & \cite{nguyen2019scalable} \\ 
DevNet & Uses supervised learning to estimate deviations of data points from normal behavior by directly optimizing a deviation loss function. & Neural Network & 2019 & \cite{pang2019deep} \\ 
LUNAR & Integrates reconstruction-based methods with self-supervised learning to detect anomalies in datasets lacking labeled samples. & Reconstruction-based NN & 2022 & \cite{goodge2022lunar} \\ 
\bottomrule
\end{tabular}
}
\vspace{-0.1in}
\label{table:algorithms}
\end{table*}

\subsection{Automated Model Selection Pipeline}
\label{subsec:llm}

We summarize the pipeline in Fig.~\ref{fig:overview}. 
In this three-step process, the system first extracts symbolic metadata from each model to represent its strengths and weaknesses. 
It then profiles the dataset to produce a set of symbolic tags describing its statistical characteristics. 
Finally, it compares these model and dataset tags, ranking candidate models and applying LLM-based reasoning to choose the most suitable one. 
We now discuss each step in detail.

\noindent \textbf{Step 1: Encoding Symbolic Metadata for Models.}
The first step encodes each OD model's core properties as structured symbolic metadata. 
By analyzing model papers and source code, we extract details regarding their strengths and weaknesses, storing them as:
\begin{equation}
    \mathcal{M}_\text{meta}(m_i) = \{\text{strengths}(m_i), \text{weaknesses}(m_i)\},
\end{equation}
where each $m_i \in \mathcal{M}$ is associated with strengths that highlight its suitability for certain conditions (e.g., ``effective in high-dimensional data") and weaknesses indicating scenarios where it may falter (e.g., ``computationally heavy"). These symbolic descriptions establish a consistent basis for subsequent reasoning about model selection.

\noindent \textbf{Step 2: Generating Dataset Profiles and Symbolic Tags.}
Next, we profile each dataset $\mathcal{D}$ by computing descriptive statistics that characterize its complexity and distribution:



\begin{equation}
\begin{aligned}
    \mathcal{A}_\mathcal{D} = \big\{ &\text{dimensionality}(\mathcal{D}), \text{skewness}(\mathcal{D}),\\
    &\text{kurtosis}(\mathcal{D}), \text{noise level}(\mathcal{D}), \dots\big\}.
\end{aligned}
\end{equation}

These attributes are then mapped into a set of symbolic tags:
\begin{equation}
\mathcal{T}_\mathcal{D} = \psi_{\text{LLM}}(\mathcal{A}_\mathcal{D}),
\end{equation}
where $\psi_{\text{LLM}}(\cdot)$ leverages LLMs to convert raw metrics into standardized tags (e.g., ``imbalanced data" or ``noisy features"). This transformation enables direct comparison between dataset properties and model metadata.

\noindent \textbf{Step 3: Model Selection via Symbolic-Neural Reasoning.}
In the final step, the pipeline applies symbolic reasoning and LLM-guided refinement to select a suitable model. 
Each model $m_i$ is evaluated by comparing its strengths and weaknesses against $\mathcal{T}_\mathcal{D}$:


\begin{equation}
\begin{aligned}
S(m_i, \mathcal{T}_\mathcal{D}) = \text{sim}\big(&\text{strengths}(m_i), \mathcal{T}_\mathcal{D}\big)\\
- &\text{penalty}\big(\text{weaknesses}(m_i), \mathcal{T}_\mathcal{D}\big),
\end{aligned}
\end{equation}

where $\text{sim}(\cdot)$ measures how well the model's strengths align with dataset tags, and $\text{penalty}(\cdot)$ quantifies the impact of misaligned weaknesses.

Models with scores $S(m_i, \mathcal{T}_\mathcal{D})$ above a chosen threshold $\delta$  form a candidate set (i.e., smaller $\delta$ leads to more candidates):
\begin{equation}
    \mathcal{M}^* = \big\{m_k \mid S(m_k, \mathcal{T}_\mathcal{D}) \geq \delta, \, m_k \in \mathcal{M}\big\}.
\end{equation}
The LLM then refines the choice among these top candidates by considering subtler trade-offs and contextual cues, selecting:
\begin{equation}
m^* = \arg\max_{m_k \in \mathcal{M}^*} \psi_{\text{LLM}}(m_k, \mathcal{T}_\mathcal{D}).
\end{equation}
This structured and adaptive reasoning ensures that the final selected model aligns with the dataset's characteristics and leverages the LLM's capability to incorporate additional context beyond simple numeric matching.

\noindent
\textbf{Key Advantages}. 
First, the symbolic reasoning step enhances interpretability by explicitly documenting the rationale behind model selection, ensuring transparency in the decision-making process. Second, the integration of neural refinement introduces flexibility, allowing the pipeline to effectively handle complex or poorly defined datasets that may lack clear structure. 
Lastly, the automated nature of the pipeline minimizes the need for manual intervention, making it highly scalable across diverse datasets and easily adaptable to incorporate new models as they emerge.




\section{Demo of \ours}
\label{sec:demo}

\noindent
\textbf{Case 1: Outlier Detection in Five Lines of Code}.
This example shows how to train and evaluate a LUNAR model (Learning Universal Normal Abnormality Representations) with minimal code. The LUNAR model is one of the deep learning-based outlier detection algorithms included in \ours. By integrating it into an existing PyOD workflow, users can quickly assess anomalies in their data.

\begin{lstlisting}
# Example: training a LUNAR detector.
from pyod.models.lunar import LUNAR
clf = LUNAR()
clf.fit(X_train)
# Outlier scores for training data.
y_train_scores = clf.decision_scores_ 
# Outlier scores for test data.
y_test_scores = clf.decision_function(X_test)
\end{lstlisting}
With these few lines, the model is trained, and outlier scores can be easily obtained for both training and test datasets. Refer to the PyOD documentation for further examples and customization options.

\noindent
\textbf{Case 2: Automated Model Selection Using LLMs}. 
In this example, we demonstrate the automated selection and training of an outlier detection model using a LLM-powered framework. 
The \texttt{AutoModelSelector} class identifies the most suitable model from \ours' extensive collection. It evaluates the dataset's statistical properties and uses symbolic-neural reasoning to produce a recommended model. By reducing the need for manual tuning, this approach is especially valuable for complex or unfamiliar datasets.

An optional parameter \texttt{additional\_notes} allows users to provide domain insights about the dataset. 
Although these notes can guide the selection, the method can perform well using purely statistical attributes. After selecting a model,  \texttt{get\_top\_clf()} returns the best candidate, which can then be trained and evaluated.


\begin{lstlisting}
# Additional note: optional comments on the dataset.
selector = AutoModelSelector(
        dataset=X_train, 
        api_key=api_key, 
        additional_notes=additional_notes
)

# Get the selected model and the associated reason.
selected_model,reason = selector.model_auto_select()

# Initialize the top model.
clf = selector.get_top_clf()

# Train the model.
clf.fit(X_train)

# Outlier scores for training data.
y_train_scores = clf.decision_scores_

# Outlier scores for test data.
y_test_scores = clf.decision_function(X_test)
\end{lstlisting}

\newpage
\section{Experiment Settings and Results}

\noindent
\textbf{Datasets and Models}.
We evaluated our approach on a diverse collection of 17 common datasets obtained from ADBench \cite{han2022adbench}: pima, cardio, mnist, arrhythmia, pendigits, shuttle, letter, musk, vowels, optdigits, satellite, lympho, ionosphere, wbc, glass, satimage-2, and vertebral. 
For each dataset, we consider 10 deep anomaly detection models from Table \ref{tab:dl_models}: MO-GAAL, AutoEncoder, SO-GAAL, VAE, AnoGAN, DeepSVDD, ALAD, AE1SVM, DevNet, and LUNAR. 
Our objective is to select a single model from these candidates for each dataset, thereby demonstrating how the \texttt{AutoModelSelector} can identify the most suitable method given the dataset's characteristics.


\noindent
\textbf{Baselines and Evaluation}.
We compare five approaches: (1) the average performance of all models, which simulates random user choices; (2) an AutoEncoder and (3) LUNAR, representing a common baseline and a strong standalone deep learning method; and (4) two versions of the \texttt{AutoModelSelector}, one with and one without additional notes. We evaluate these methods based on their rankings regarding the area under the ROC curve (AUROC) across multiple datasets. Lower ranking values indicate better performance.

\begin{figure}[h]
    \centering
    \includegraphics[width=1\linewidth]{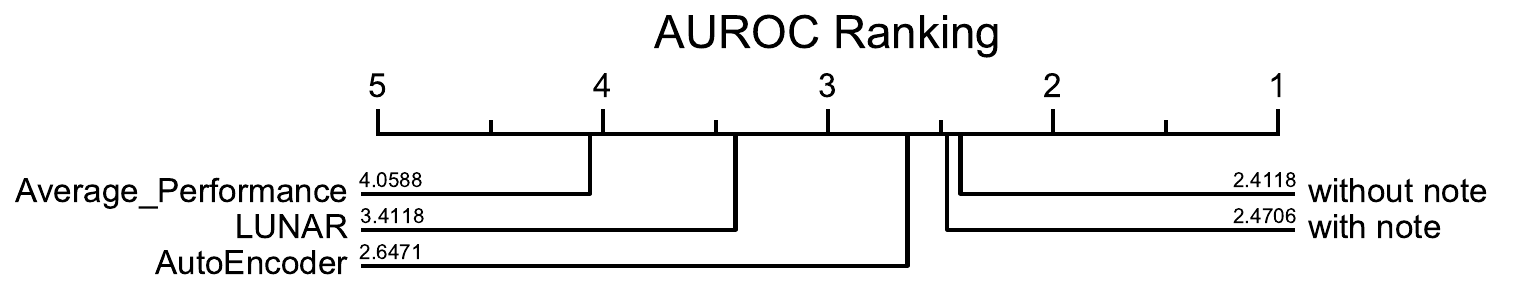}
    \caption{Comparison of AUROC rankings of baseline methods, the \texttt{AutoModelSelector} (with and without additional notes), and two standalone models (AutoEncoder and LUNAR). Lower ranks indicate better performance.}
    \vspace{-0.2in}
    \label{fig:cd_diagram}
\end{figure}

Figure~\ref{fig:cd_diagram} presents the AUROC rankings for all evaluated approaches. 
Our \texttt{AutoModelSelector} without additional notes achieves the \textbf{best} ranking (2.4118), slightly surpassing even the variant that leverages user-provided notes (2.4706). 
While domain insights can be helpful, these results indicate that the selector's inherent reasoning capabilities can guide effective model selection, even in the absence of extra contextual details.

In contrast, the standalone AutoEncoder model attains a ranking of 2.6471 and LUNAR scores 3.4118, both trailing behind the performance of the \texttt{AutoModelSelector}. 
The average performance of all models is the weakest at 4.0588, illustrating that adaptive, data-driven selection methods significantly outperform naive strategies. 
Collectively, these findings highlight the robust effectiveness of our automated selection framework, which identifies well-suited models and delivers strong OD performance across diverse datasets.

\section{Conclusion and Future Directions}

We introduced \ours, a significant update to the PyOD library, emphasizing integrating state-of-the-art deep learning models under a unified PyTorch framework and introducing a large language model (LLM)-driven automated model selection process. 
By offering an expanded set of 45 outlier detection algorithms, including 12 recently developed deep learning methods, \ours streamlines the deployment and evaluation of outlier detection systems. These enhancements support both researchers and practitioners by simplifying model integration, improving accessibility, and reducing the barriers to identifying suitable methods for specific datasets.

Several opportunities remain for further development. 
One direction is to explore the integration of domain-specific priors, leveraging knowledge from fields such as cybersecurity, healthcare, and online retail to guide automated model selection more efficiently. 
Another direction involves extending the framework to support continual learning and adaptive pipelines that adjust to shifting data distributions. Expanding the LLM-based reasoning module to incorporate advanced feedback loops and user interaction could also refine the decision-making process. 
Additionally, there is potential to improve computational scalability and incorporate distributed training strategies to handle larger and more complex datasets. 
\bibliographystyle{ACM-Reference-Format}

\bibliography{reference,sample-base}


\end{document}